%% file: 0main.tex
\documentclass{article}
\usepackage[sglblindworkshop, final]{neurips_2026}
\workshoptitle{Interpreting Agent Behavior (IAB) -- GLEE Competition Track}
\usepackage[utf8]{inputenc}
\usepackage[T1]{fontenc}
\usepackage{array,booktabs,float,microtype,amsmath,amssymb,graphicx,tikz}
\usepackage{enumitem}
\usepackage{placeins}
\usepackage[hidelinks]{hyperref}
\usepackage{url}
\usetikzlibrary{arrows.meta,positioning,fit}

\newcommand{\code}[1]{\texttt{#1}}
\title{Self-Evolving Agentic Workflows\\via Semantic Decomposition of Complex Tasks}
\author{%
  Zichuan Fu$^{1}$\quad Junjia Qi$^{1}$\quad Xian Wu$^{2}$\quad Wenlin Zhang$^{1}$\\
  \textbf{Yimin Deng$^{1}$\quad Kaifeng Guo$^{1}$\quad Xiangyu Zhao$^{1}$}\thanks{Corresponding author.}\\
  $^{1}$City University of Hong Kong\quad $^{2}$Tencent Jarvis Lab\\
  \texttt{zc.fu@my.cityu.edu.hk}, \texttt{xianzhao@cityu.edu.hk}
}
\begin{document}
\maketitle
\begin{abstract}
Agentic workflows organize complex tasks into reusable procedures. However, invoking an LLM at every step incurs repeated inference costs and delays. We propose a framework for self-evolving agentic workflows through semantic task decomposition. Each subtask has explicit applicability conditions and a procedure that executes code or invokes a model as needed. An agent uses execution feedback to refine the decomposition and update these procedures. Experiments in the GLEE competition show the framework on bargaining, negotiation and persuasion. Most observed final-decision procedures are implemented in code. Code is available at \href{https://github.com/Applied-Machine-Learning-Lab/NeurIPS2026_SeaFlow}{Applied-Machine-Learning-Lab/NeurIPS2026\_SeaFlow}.
\end{abstract}
\input{1Introduction}
\input{2RelatedWork}
\input{3Method}
\input{4Experiments}
\input{5BehaviorAnalysis}
\FloatBarrier
\input{7LimitationsConclusion}
\clearpage
\bibliography{references}
\bibliographystyle{plainnat}
\appendix
\input{8Appendix}
\input{9ExtendedExperiments}
\end{document}

%% file: 1Introduction.tex
\section{Introduction}
LLM-based agents use model reasoning to carry out complex tasks. Agentic workflows organize this process into explicit steps whose connections determine execution~\citep{khattab2024dspy,zhang2025aflow}. The resulting procedure can be reused across inputs. An agent can construct the procedure and revise it using execution feedback~\citep{hu2025adas,cheng2024trace}.

When every workflow step requires a model call, inference overhead grows with the number of steps, on top of the reasoning cost within each call~\citep{fu2026tandem}. The resulting delay can make a workflow unsuitable for time-sensitive interactions, while repeated execution increases its cost. Some of this overhead comes from using model reasoning for steps that already have defined solutions. Recent work reduces this dependence by making more of the decision process directly executable~\citep{xu2026hyevo,zhang2026copitree,qi2026llmascode}. This motivates task decompositions that specify where agent reasoning is needed and leave the remaining steps to code.

Our framework begins with (1) semantic task decomposition, through which an agent organizes decisions into subtasks based on the conditions observed during interaction. This decomposition provides the basis for (2) defining a solution procedure for each subtask, so model reasoning need not be used throughout the workflow. Execution then provides evidence for (3) revising the workflow. When a subtask handles an observed situation poorly, the agent changes its scope or solution procedure. We examine this process across development iterations in three GLEE environments~\citep{shapira2024glee}, using recorded actions to show how the resulting workflows make decisions.

\input{1Contributions}

\begin{figure}[!t]
\centering
\input{figures/method_overview}
\caption{(a) Bargaining decision paths at revision \texttt{a1875a0}; leaves return plan parameters. (b) The agent's workflow revision cycle, from execution feedback to a retained workflow.}
\label{fig:method}
\end{figure}
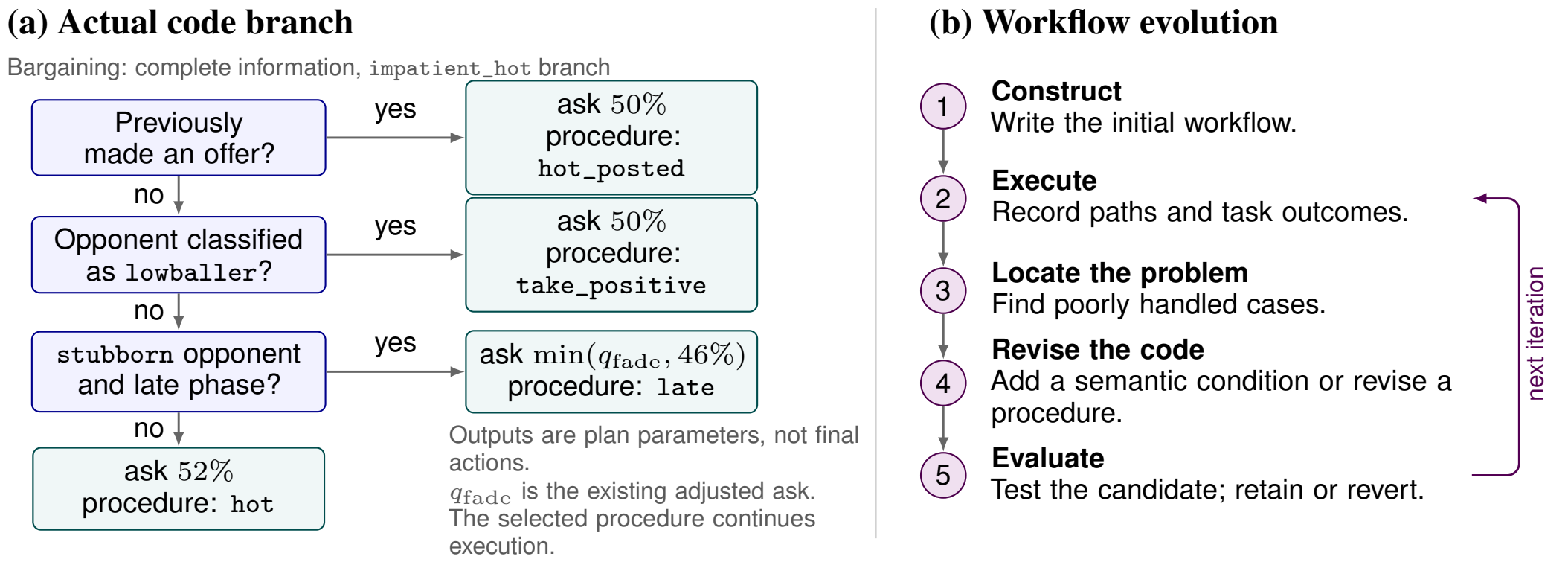

%% file: 1Contributions.tex
Our main contributions are as follows:
\begin{itemize}[leftmargin=*, itemsep=0pt]
\item We propose a framework for constructing agentic workflows through semantic task decomposition. The framework organizes complex interactive decisions into a hierarchy of subtasks, with each subtask defining when it applies and how it is solved.
\item We develop a feedback-driven method for improving these workflows. An agent uses execution records and task results to identify problems. It then divides subtasks further or changes how they are solved in later iterations of the evolving workflow.
\item We evaluate the method across three GLEE environments and analyze 725 development iterations and 12,249 action records collected with fixed workflows. We examine how strategies emerge, change, or are removed, and how different parts of the workflow contribute to decisions over time.
\end{itemize}

%% file: figures/method_overview.tex
\begin{tikzpicture}[
 x=1cm,y=1cm,font=\sffamily\scriptsize,
 edge/.style={-{Latex[length=1.3mm]},draw=black!60,line width=.55pt},
 test/.style={draw=blue!55!black,fill=blue!5,rounded corners=2pt,
 text width=2.4cm,minimum height=.58cm,align=center,inner sep=3pt},
 leaf/.style={draw=teal!60!black,fill=teal!6,rounded corners=2pt,
 text width=2.38cm,minimum height=.58cm,align=center,inner sep=3pt},
 number/.style={circle,fill=violet!12,draw=violet!60!black,minimum size=.40cm,inner sep=1pt},
 evolutiontext/.style={align=left,text width=4.1cm,anchor=west,inner sep=0pt}
]
\path[use as bounding box] (0,.3) rectangle (13.8,-4.6);
\node[font=\small\bfseries,anchor=west] at (.05,.1) {(a) Actual code branch};
\node[font=\small\bfseries,anchor=west] at (8.25,.1) {(b) Workflow evolution};
\draw[black!18] (7.90,.24)--(7.90,-4.47);
\node[anchor=west,text=black!65,font=\sffamily\tiny] at (.05,-.30)
 {Bargaining: complete information, \texttt{impatient\_hot} branch};
\node[test] (posted) at (1.70,-.91) {Previously made an offer?};
\node[leaf] (postedplan) at (5.55,-.91) {ask $50\%$\\procedure: \texttt{hot\_posted}};
\node[test] (low) at (1.70,-1.95) {Opponent classified\\as \texttt{lowballer}?};
\node[leaf] (lowplan) at (5.55,-1.95) {ask $50\%$\\procedure: \texttt{take\_positive}};
\node[test] (stubborn) at (1.70,-2.99) {\texttt{stubborn} opponent\\and late phase?};
\node[leaf] (lateplan) at (5.55,-2.99) {ask $\min(q_{\rm fade},46\%)$\\procedure: \texttt{late}};
\node[leaf] (defaultplan) at (1.70,-4.03) {ask $52\%$\\procedure: \texttt{hot}};
\draw[edge] (posted)--node[above]{yes}(postedplan);
\draw[edge] (posted)--node[left]{no}(low);
\draw[edge] (low)--node[above]{yes}(lowplan);
\draw[edge] (low)--node[left]{no}(stubborn);
\draw[edge] (stubborn)--node[above]{yes}(lateplan);
\draw[edge] (stubborn)--node[left]{no}(defaultplan);
\node[align=left,text width=3.65cm,anchor=west,font=\sffamily\tiny,text=black!65]
 at (3.98,-4.04)
 {Outputs are plan parameters, not final actions.\\$q_{\rm fade}$ is the existing adjusted ask.\\The selected procedure continues execution.};

\node[number] (n1) at (8.50,-.63) {1};
\node[evolutiontext] at (8.92,-.63) {\textbf{Construct}\\Write the initial workflow.};
\node[number] (n2) at (8.50,-1.45) {2};
\node[evolutiontext] at (8.92,-1.45) {\textbf{Execute}\\Record paths and task outcomes.};
\node[number] (n3) at (8.50,-2.27) {3};
\node[evolutiontext] at (8.92,-2.27) {\textbf{Locate the problem}\\Find poorly handled cases.};
\node[number] (n4) at (8.50,-3.09) {4};
\node[evolutiontext] at (8.92,-3.09) {\textbf{Revise the code}\\Add a semantic condition or revise a procedure.};
\node[number] (n5) at (8.50,-3.91) {5};
\node[evolutiontext] at (8.92,-3.91) {\textbf{Evaluate}\\Test the candidate; retain or revert.};
\draw[edge] (n1)--(n2);
\draw[edge] (n2)--(n3);
\draw[edge] (n3)--(n4);
\draw[edge] (n4)--(n5);
\draw[edge,draw=violet!65!black,rounded corners=3pt]
 (13.2,-3.91)--(13.62,-3.91)--(13.62,-1.45)--(13.2,-1.45);
\node[rotate=90,font=\sffamily\tiny,text=violet!65!black] at (13.76,-2.65) {next iteration};
\end{tikzpicture}

%% file: 2RelatedWork.tex
\section{Related Work}

\paragraph{Agentic workflows organize steps by interface or role.}
Workflow representations organize computational steps and their connections. DSPy composes declarative model-call modules~\citep{khattab2024dspy}. SEW connects role-based agents whose tasks are specified by prompts~\citep{liu2025sew}. AFlow writes model calls and operators as code~\citep{zhang2025aflow}. HyEvo combines language model and code nodes~\citep{xu2026hyevo}. LLM-as-Code keeps control flow in code and calls models only for reasoning~\citep{qi2026llmascode}. These designs organize a workflow around its interfaces and roles. The situation that each step serves stays implicit in this structure. Our workflows name these situations as explicit conditions and attach a procedure to each one.

\paragraph{Self-evolving agents generate or revise programs.}
Agent improvement ranges from candidate search to targeted revision. TRACE updates program components from execution traces and feedback~\citep{cheng2024trace}. ADAS generates programs from an archive of evaluated agents~\citep{hu2025adas}. LLEGO guides tree crossover and mutation with semantic priors~\citep{liu2025llego}. EvoSkillRec evolves recommender architectures by recombining validated skills and generating new ones~\citep{li2026evoskillrec}. Co-$\pi$-tree revises problematic branches~\citep{zhang2026copitree}. Trace2Policy patches rules from grouped errors~\citep{zha2026trace2policy}. HexMachina refines policies through simulation~\citep{belle2025hexmachina}. These methods generate or revise programs, and the situations each part handles remain implicit. Our agent names these situations and revises their boundaries, so a revision can split or merge the game situations that one subtask handles.

%% file: 3Method.tex
\section{Self-Evolving Agentic Workflows}
Our method separates the workflow that makes decisions from the agent that improves it. The workflow is executable code and remains fixed during an interaction. After execution, the agent may revise that code using the resulting records and task outcomes (Figure~\ref{fig:method}).

\subsection{Programmatic Workflow}
At iteration $k$, workflow $W_k$ specifies a decision structure with its parameters and solution procedures. Given input $x_t$ and interaction history $h_t$, it returns an action $a_t$ and execution trace $\tau_t$:
\begin{equation}
    (a_t,\tau_t)=W_k(x_t,h_t)
    \label{eq:workflow}
\end{equation}
The program may use conditions, function calls, nesting, and loops, so its structure is not restricted to a decision tree (Appendix~\ref{app:structure}). Each operation can execute code directly or invoke a model. A model operation may use an LLM for unresolved language-dependent decisions or a learned predictor for an estimated quantity. The surrounding program determines when an operation is used, supplies its inputs, and processes its output. Model inference occurs only where specified by the workflow.

\subsection{Workflow Evolution}
At iteration $k$, the agent executes $W_k$ and collects execution traces and task outcomes $D_k$. It examines poorly handled cases to find problems in task decomposition, parameters or solution procedures. It then constructs a candidate workflow from $W_k$ and $D_k$
\begin{equation}
    \widetilde W_{k+1}=\mathcal{A}(W_k,D_k)
    \label{eq:evolution}
\end{equation}
Here, $\mathcal{A}$ selects revisions according to the diagnosed problem.
\emph{(1) Task decomposition:} split subtasks through observable conditions or reorganize branches when distinct cases require different decisions.
\emph{(2) Parameter adjustment:} tune thresholds, weights, or other numerical settings within an otherwise suitable procedure.
\emph{(3) Model fitting:} train or update classification or regression models from execution data to estimate acceptance probabilities or continuation values.
\emph{(4) Procedure redesign:} devise an alternative strategy and implement it through code or model reasoning when the current solution is inadequate.
The agent may combine these operations within an iteration or reconsider earlier choices as further execution provides new evidence.

The candidate is evaluated using replay, held-out records, or bounded deployment to assess the targeted behavior and check for regressions (Appendix~\ref{app:revision}). If retained, it becomes $W_{k+1}$; otherwise, $W_{k+1}=W_k$. The next iteration executes the retained workflow and collects new feedback. Revisions occur between interactions, while the workflow remains fixed during each interaction.

%% file: 4Experiments.tex
\section{Experiments}
We evaluate competition performance (RQ1), workflow evolution (RQ2) and agent behavior (RQ3).

\subsection{Experimental Settings}
We evaluate our agent on Bargaining, Negotiation and Persuasion in GLEE against the other competition agents under the competition rules (Appendix~\ref{app:compliance}). Performance is the official GLEE rating, and efficiency is the local decision time per action. GPT-5.5 revises the workflows from execution feedback via our OpenProgram framework\footnote{\url{https://github.com/fzkuji-neo/OpenProgram}}, and Grok-4.6 serves the optional runtime LLM modules. Workflows stay fixed within a game. Appendix~\ref{app:logs} describes the logs and commit histories.

\subsection{Competition Results (RQ1)}
This section presents the overall performance of the final workflows. Our agent ranks first in the GLEE competition, and its ratings in all three environments rise as the workflows evolve (Table~\ref{tab:competition}, Figure~\ref{fig:strategy-timeline}). In all three environments, the final workflows make each decision locally within a millisecond. Appendix~\ref{app:results} gives the results, samples and strategies of each game.

\begin{table}[!t]
\caption{Competition ratings and mean statistics of the final workflows (revision \texttt{a1875a0}).}
\label{tab:competition}
\centering
\begin{tabular}{@{}lrrr@{}}
\toprule
Measure & Bargaining & Negotiation & Persuasion \\
\midrule
Official rating $\uparrow$ & 2760.27 & 2761.67 & 2261.58 \\
\midrule
Mean branches per strategy node & 2.13 & 2.25 & 2.24 \\
Mean strategy judgments per action & 37.61 & 22.95 & 7.42 \\
Mean local decision time (ms) & 0.118 & 0.198 & 0.122 \\
\bottomrule
\end{tabular}
\end{table}

\subsection{Workflow Evolution (RQ2)}
\begin{figure}[!t]
\centering
\includegraphics[width=\textwidth]{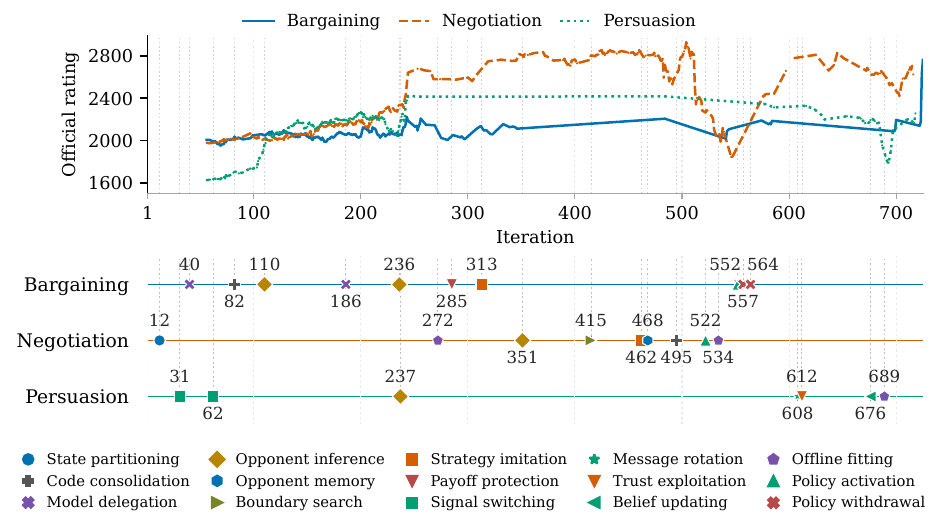}
\caption{Competition rating history and strategy changes on the same 725-iteration axis. Top: recorded deployment segments, one rating per plotted iteration. Bottom: representative strategy changes, with iteration numbers and operation markers.}
\label{fig:experiments}
\label{fig:strategy-timeline}
\end{figure}

The agent keeps revising the workflow strategies throughout development. The 26 representative strategy changes in Figure~\ref{fig:strategy-timeline} span iterations 12 to 689, and four of them come after iteration 600. Revisions withdraw strategies as well as add them (Appendix~\ref{app:revision}). A model judge enters Bargaining at iteration 40 and is disabled at 186. The plateau exit enters at 552 and is withdrawn at 564, while a separate Bargaining responder policy returns to shadow mode at iteration 557.

Each environment follows its own path, since every revision answers the records of its own games. Negotiation first separates roles and hidden types at iteration 12. It later probes the acceptance interval (415) and mirrors seller concessions (462). It also remembers opponents across games (468). Persuasion switches its signals at 31 and 62 and sells a low-quality product after trust is built (612). Bargaining protects its payoff at 285 and mirrors revealed demands at 313. The initial workflows contain none of these procedures, which the agent derives from game records.

Ratings move while the workflows change, and the shared axis places each change against them. All three environments rise together around iterations 236--244. The figure shows continuous revision. It does not isolate the effect of any single change on the rating.

The competition rating rewards only a player's own payoff. It ranks this payoff among players in the same configuration and role, adjusted only for the opponent's rating. Neither the opponent's payoff nor the joint payoff enters the rating, and a deal that enlarges the total earns no extra credit. Our evolved workflows become firm in asking and rejecting. The final Bargaining workflow opens at a 55\% self-share in all 63 eligible states. Its reservation lock rejects in 382 states. From iteration 285 it also rejects non-final offers that leave it at most 10\%.

%% file: 5BehaviorAnalysis.tex
\subsection{Agent Behavior Analysis (RQ3)}
\begin{figure}[!t]
\centering
\begin{minipage}[t]{0.49\textwidth}
\centering
\includegraphics[width=\linewidth]{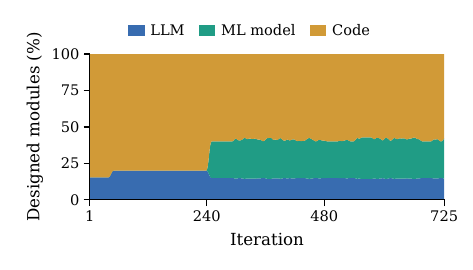}
\par\small (a) Functional modules
\end{minipage}\hfill
\begin{minipage}[t]{0.49\textwidth}
\centering
\includegraphics[width=\linewidth]{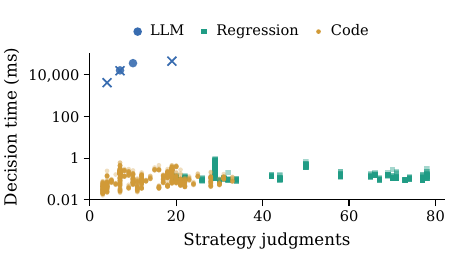}
\par\small (b) Strategy judgments and latency
\end{minipage}
\caption{(a) Shares of designed decision modules by executor across iterations. (b) Decision time per executor against strategy judgments per decision (log scale). Crosses mark failed LLM calls.}
\label{fig:decision-profile}
\end{figure}

Our agent decides through three kinds of modules, namely code, fitted machine-learning models and LLM calls. Code makes up most of the designed modules throughout development (Figure~\ref{fig:decision-profile}a). Fitted models join as the workflows learn from recorded games, and LLM modules remain as optional interfaces. These shares count designed modules rather than decisions (Appendix~\ref{app:hybrid}). Recorded actions follow the coded conditions, as with the opening rule in every eligible Bargaining state.

Code and fitted models decide in under a millisecond, and the measured LLM calls take seconds and mostly fail (Figure~\ref{fig:decision-profile}b). Decision time depends on the executor and hardly on the number of strategy judgments. The fast executors let the agent make every decision locally within its turn.

\begin{figure}[!t]
\centering
\includegraphics[width=\textwidth]{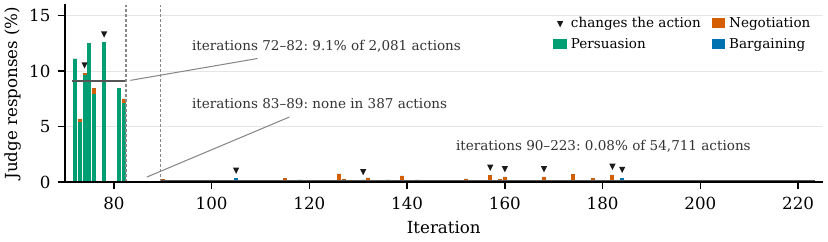}
\caption{Share of non-message actions with a valid LLM-judge response per iteration, by task. Triangles mark responses that change the action. Dotted lines mark Persuasion leaving the judge.}
\label{fig:llm-share}
\end{figure}

In actual play, the LLM judge decides rarely and less over time (Figure~\ref{fig:llm-share}). Early on it answers about one in eleven non-message actions, almost all of them Persuasion buyer decisions. Its answers agree with the code in all but one case. The agent then makes these decisions deterministic, and judge responses almost vanish after iteration 90. Code makes nearly all final decisions from then on.

%% file: 7LimitationsConclusion.tex
\section{Conclusion}
We present self-evolving agentic workflows that decompose tasks into executable subtasks and revise their procedures from interaction feedback. The GLEE study documents structural and procedural revisions, with submillisecond mean local decision times. Controlled comparisons are needed to quantify performance gains and costs across workflow construction, execution, and revision.

%% file: 8Appendix.tex
\section{Structure of the Final Workflows}
\label{app:structure}
\label{app:details}
This section shows what the final workflows look like. It gives the order in which the code reaches one move and the procedures at each game's leaves. It then reports which branches the recorded states actually reach.

\subsection{Execution Order for One Move}
\begin{figure}[tbp]
\centering
\includegraphics[width=\textwidth]{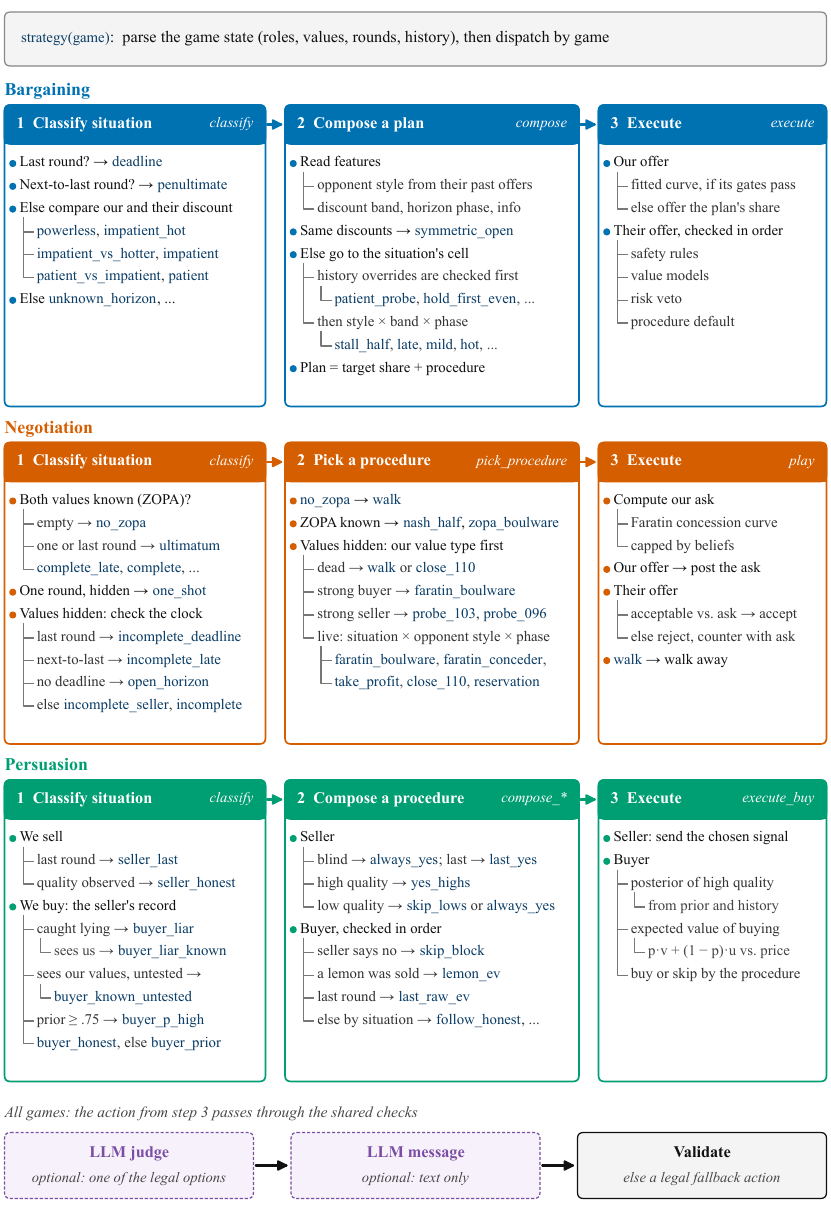}
\caption{Execution order of the final workflows for one move. Within a box, rows are checked from top to bottom and indentation marks nested checks; dashed boxes are optional model calls.}
\label{fig:decision-logic}
\end{figure}

The final code reaches each move through three levels, and each level answers a different question (Figure~\ref{fig:decision-logic}). After the game is identified, ordered rules first name the current situation. A second step reads further features and selects a procedure for that situation. A last step turns the procedure into a concrete offer or decision. The second level differs across games because their main uncertainty differs. Bargaining reads the opponent's concession style from past offers. Negotiation first reads the type of our own hidden value, which decides whether a profitable deal can exist. Persuasion separates the seller and buyer roles and then tracks how credible the seller's signals have been. Keeping these questions on separate levels lets a revision change one node while its siblings keep their behavior, as the revision protocol in Appendix~\ref{app:protocol} requires. All numbers come from code. The optional model calls at the end can only choose among precomputed legal options or write message text, which is consistent with the rare model decisions in Appendix~\ref{app:hybrid}.

\subsection{Leaf Procedures by Game}
In Bargaining, regime, opponent style, inflation, horizon, and information select a named plan. Offer turns may use an evidence-gated fitted offer before legal construction; response turns apply safety rules, planned-continuation comparisons, fitted and trajectory models, risk vetoes, and procedure-specific fallbacks in precedence order. In Negotiation, the workflow computes own profit before opponent inference, separates complete from incomplete information and open from finite horizons, and applies ZOPA, Faratin/Boulware, take-profit, probe, reservation, or walk procedures. In Persuasion, signals and recommendation history update buyer belief and seller credibility; leaves implement expected-value tests and recommendation rules.

\subsection{Bargaining Offer Model}
The Bargaining offer fit standardizes share, progress, both visible discounts, rejection count, last opponent share, and role. A logistic model estimates acceptance and a ridge model estimates payoff after rejection; the leaf maximizes discounted expected value over 35--60\% shares plus the rule fallback. Training requires minimum effective sample sizes, chronological holdout improvement over a constant baseline, and a negative share coefficient for acceptance. Runtime replacement additionally requires an estimated edge above .015. The current serialized accept and continuation tables have no shared segment, so offer fitting returns the rule fallback. Response fitting instead learns segment-specific acceptance margins under train, holdout, and external replay gates. Serialized weights are loaded once per process and remain fixed during a deployment.

\subsection{Branch Usage on Replayed States}
\begin{figure}[tbp]
\centering
\includegraphics[width=\textwidth]{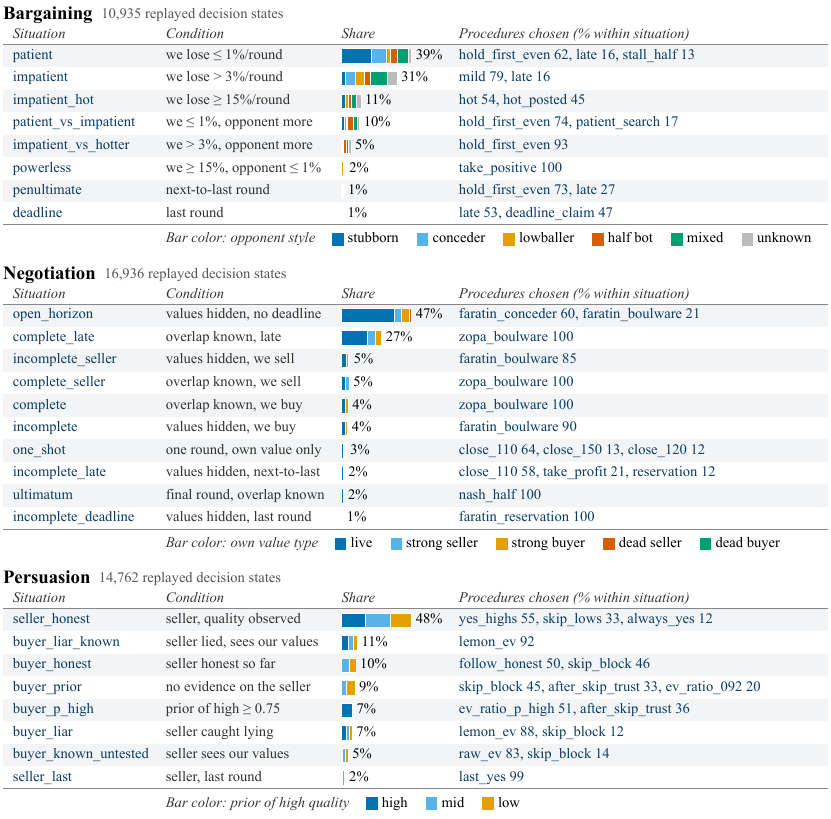}
\caption{Situations and procedures of the final workflows on replayed decision states. Each row is one situation; its bar gives the share of states, colored by the second-level split, and the right column lists the procedures chosen within that situation.}
\label{fig:decision-hierarchy}
\end{figure}

A tree with many branches helps only if its branches handle real states, so we also check where decisions go. We replay the decision states recorded during development through the final revision. For each state, we log its situation, its second-level split, and the procedure it reaches (Figure~\ref{fig:decision-hierarchy}). In Bargaining and Negotiation, two situations receive most states; in Persuasion, the seller who observes product quality receives about half. Within a situation, one or two procedures handle most states. The second level matters most in the large situations: the patient Bargaining situation, for instance, meets every opponent style. Rare situations, such as the last round or one-round games, still keep their own leaves, which were added for edge cases met during development. The second-level splits also show how the workflows read their opponents. In about a third of Persuasion buyer states, the seller is classified as lying. These states mostly value the product by its expected value under a lemons prior. Bargaining meets stubborn and mixed concession styles most often.

These shares come from replayed development states rather than a new tournament, so they reflect the opponents met during development. They also show only which path the code takes. They do not measure how much each branch adds to the rating or whether its decisions are optimal.

\subsection{Visited Decision Units}
We deduplicate final selection node IDs in the Bargaining action traces and final procedure names in Negotiation and Persuasion. This yields 58, 15, and 16 observed units, respectively. The inventory counts only visited final-decision units, not all program branches, intermediate computations, or dormant LLM call sites. Under the existing final-executor taxonomy, Bargaining has 53 code units, one predictor unit, and four unclassified units. Negotiation has 14 code units and one procedure whose nine fitted-eligible actions do not reveal whether the predictor displaced its numerical fallback. Persuasion has 16 code units. Code combines rule and numerical labels; it does not imply that no model was used in an upstream calculation. Runtime LLM judging is disabled in these samples. The exported inventory records each unit's identifier, category, and action count; node cardinalities and execution frequencies therefore remain distinct quantities.

\section{Workflow Revisions by the Coding Agent}
\label{app:revision}
This section follows the coding agent as it revises the workflows. It gives the protocol each revision follows and then inspects selected revisions in detail. It ends with the revision activity across all 725 iterations.

\subsection{Revision Protocol}
\label{app:protocol}
The recurring protocol reconstructed from repository instructions uses a shared history-inspector prompt followed by a strategy prompt and one family-specific routing prompt. The coding agent is instructed to open the exact replay, reconstruct decision-time state, compute payoffs or continuation values with diagnostic code, record the visited route, modify only the affected node, preserve sibling behavior and the parent fallback, and add a named replay. A revision names the target node, semantic distinction, executor, fallback, and validation condition; the coding agent commits the tree. The repository retains these instructions and accepted code, but not one fixed design-time model identifier or a complete proposal ledger.

\subsection{Inspected Revisions}
We inventory the changed files and functions in all 725 commits of the shared index; 530 touch the strategy-code paths considered in this inventory. We then inspect 31 family-specific changes in detail and display 26 selected events: ten in Bargaining, nine in Negotiation, and seven in Persuasion. These are evidence-backed examples, not an exhaustive classification or an estimate of how frequently each operation occurs. A commit may contain several changes; its marker identifies the specific mechanism inspected.

The mechanisms differ across environments. Negotiation introduces acceptance-boundary search, concession mirroring, and persistent opponent profiles. Persuasion instead changes which signal to send, rotates rejected message templates, and uses established trust to attempt low-quality sales. Its late buyer revision also separates two operations: updating a posterior from the current opponent's revealed outcomes and refitting prior parameters from completed games. These should not be combined into a generic adaptation category. Bargaining includes a different sequence in which candidate policies gain and subsequently lose decision authority.

Figure~\ref{fig:strategy-timeline} compares selected revisions across environments. Both panels use a single index constructed by merging the recorded main-branch, family-lineage, and additional deployment commits, deduplicating hashes, and ordering commits by timestamp, with hash order breaking ties. Each deployment rating and strategy marker is mapped by its commit hash, so their horizontal positions are directly comparable. The following case studies retain the 359-commit main-branch positions used in the source audit; these are distinct from the shared plotting index. The operation categories describe recorded changes, not prescribed development stages.

\subsection{Selected Revision Cases}
Bargaining already has discrete opponent categories in the earliest snapshot; decomposition does not first appear at a later marked commit. The subsequent history contains both more specific conditions and consolidation of existing logic. At position 82, a unified planner replaces several route-specific implementations. Positions 236--268 add continuous opponent state and fitted continuation estimates with recency weighting. We examine concrete revisions rather than treating commit-message keywords as measurements of the agent's reasoning.

\begin{figure}[htbp]
\centering
\resizebox{\textwidth}{!}{\input{figures/observed_refinement}}
\caption{An actual Bargaining revision. Left: the acceptance branch introduced at position 236, with all its conditions and the fallback to the existing procedure. Right: the motivating situation and expected behavior documented in the same commit's tests. Earlier branches retain precedence. The diagram shows a local code change, not the entire workflow.}
\label{fig:refinement}
\end{figure}
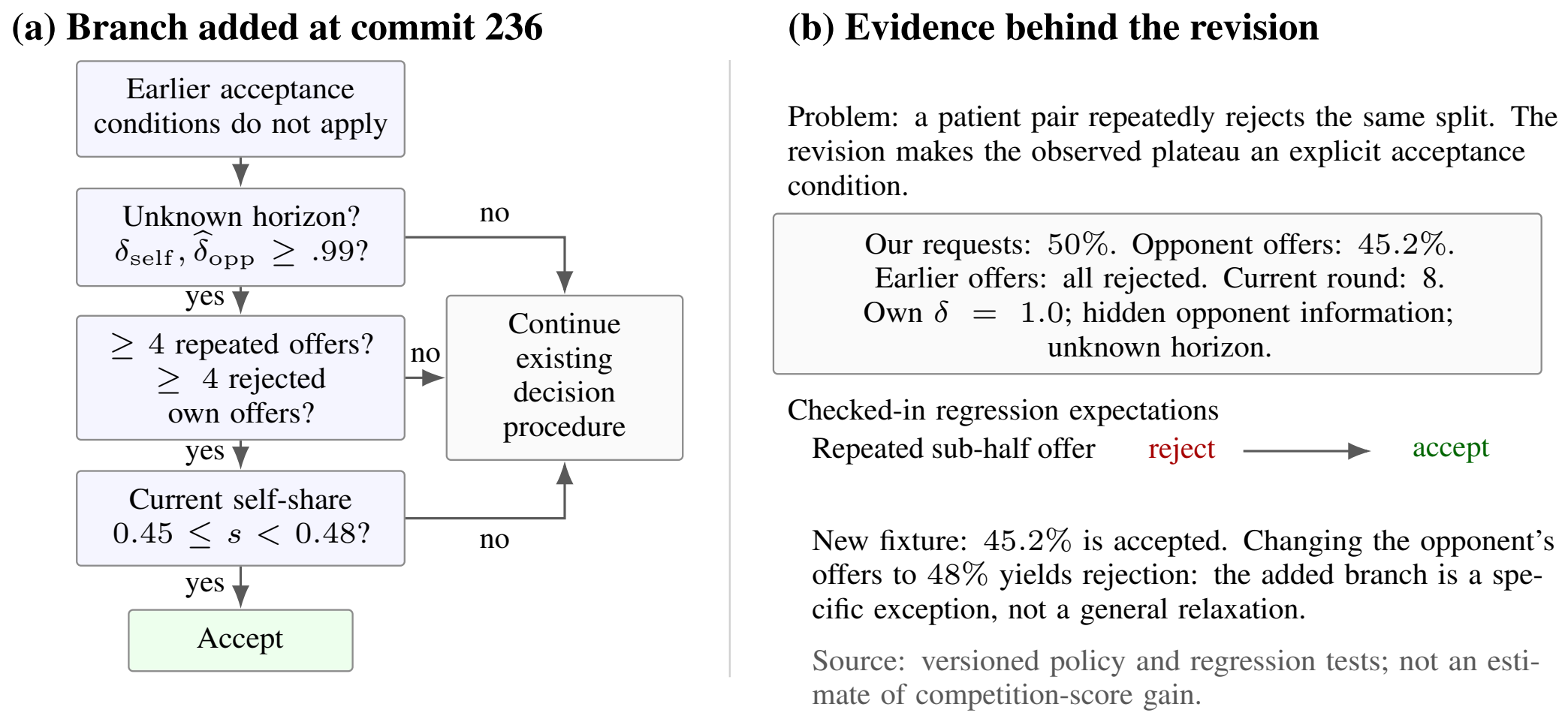

\paragraph{From an opponent category to an observed state.}
Position 236 adds an opponent-state estimator that records recent concessions, repeated offers, and rejections of our offers. The distinction affects execution: for a patient player facing an unknown horizon, a repeatedly observed offer can justify ending the interaction even below an earlier acceptance boundary. Figure~\ref{fig:refinement} reconstructs the added branch and its regression cases. Two existing tests change their expected decision from rejection to acceptance. A new test accepts a repeated $45.2\%$ self-share, while its $48\%$ variant still rejects. The change therefore implements a narrowly defined response to a plateau rather than uniformly lowering every threshold. Semantic refinement here changes which history features determine a decision, not only the wording of a prompt.

\paragraph{A proposed strategy need not remain in the workflow.}
At position 289, a candidate counteroffer mirrors the share revealed by the opponent. In its round-two, equal-discount branch, it requests a self-share of $\min(0.60,\max(0.55,1-s))$, where $s$ is the share offered to us. Position 291 removes this branch in a rollback explicitly described as withdrawing unvalidated changes. This is evidence of proposal and withdrawal, not proof that mirroring reduced scores. It also differs from opening imitation in the fixed-workflow analysis: one responds to the current opponent's offer, whereas the other applies an opening rule. Treating both as a single strategy label would conceal their different conditions and retention histories.

\paragraph{Reconsidering which evidence controls a branch.}
Position 284 removes an acceptance override triggered by phrases in the opponent's message, combined with specific share ranges. The removed branch could accept because the message announced a final offer or willingness to close. Removing it reduces the direct influence of opponent-controlled wording on that decision. Together with the state estimator, this illustrates a substantive change in policy design: deciding which observations should justify an action, rather than only adding more branches.

\paragraph{Changes outside Bargaining.}
At position 2, Negotiation separates hidden strong types from cases where a small positive gain remains acceptable; its final-round rule still accepts a nonnegative gain. The commit records earlier cases that conceded to reservation and then accepted a very small surplus. Persuasion's August~24 development record describes a different failure: skipping an early low-quality recommendation can leave an otherwise profitable buyer unwilling to buy later recommendations. The proposed response recommends low-quality items until the first sale and changes behavior thereafter. This dated inspector record supplies the development rationale, whereas the Negotiation example is supported by a versioned policy change and regression tests. These cases show why identical revision categories do not imply identical strategies across environments.

\subsection{Bargaining Milestones}
The retained Bargaining history begins with discrete semantic routing and dormant after-game adaptation. It subsequently adds anti-gifting defense at position 22, broad bounded judging at 40, ambiguity-only judging at 55, and a unified plan pipeline at 82. Bargaining judging is disabled at 186. Positions 236--268 enrich opponent state and add evidence-gated fitting, continuation modeling, bounded predictions, and recency weighting. Later changes emphasize validation: a lowball response is removed at 271, rating-risk defense appears at 281, audit instrumentation appears at 283, text-controlled routing is removed at 284, the mirroring candidate is removed at 291, and a broader response family is rolled back at 294. Position 315 introduces replay-first branch ordering. These are selected code milestones, not a complete deployment history or estimates of individual rating effects.

\subsection{Function Edit Inventory}
The separate source audit uses the 725 revisions in the shared commit index. For each revision, we compare Python function definitions in the three environment-specific workflow directories with the same files in its first parent. We count module-level functions and class methods; nested local functions remain part of their enclosing function. A definition is identified by its file path and qualified name. Additions and removals change this inventory; modifications change the abstract syntax tree of an existing definition. Formatting and comments do not count as modifications, whereas changed docstrings do. Renaming or moving a function counts as a removal and an addition. Shared workflow utilities, external model wrappers, and tests are outside this inventory. Function counts therefore measure the inspected code structure, not the number of decision nodes or model calls.

The archived source-audit data sum these edit instances across environments in trailing 50-position windows before computing proportions. A window without edits is missing rather than zero. Each version is compared with its own first parent, not with the preceding snapshot in the cross-branch plotting order. Cherry-picked changes on different branches remain separate edit instances. The function-count series includes only the 359 main-branch snapshots, mapped onto the same shared index, so lines do not connect unrelated branch states. These statistics neither identify the author of an edit nor attribute a performance gain to it.

\subsection{Revision Activity across All Iterations}
\label{app:iteration-activity}
The main text and the cases above show selected revisions. This subsection follows revision work across all 725 iterations. It asks where the code changes and which revision operation each change applies. It also asks how the coding agent justifies, undoes and checks its changes.

\paragraph{Labeling revision operations.}
We compare each iteration with its first parent at the level of functions, methods and module constants in the workflow code. Each changed unit receives one of the four revision operations of our method. A change that disappears once numeric constants are masked is a parameter adjustment. A change to the condition of an \code{if}, \code{elif} or \code{while} statement is task decomposition, since it changes when a subtask applies. Any other change to a function body is procedure redesign. Added or removed functions count as decomposition in classification and routing files and as redesign elsewhere. Changed learned artifacts and fitting code count as model fitting. Function edits follow the source audit above.

\begin{figure}[tbp]
\centering
\includegraphics[width=\textwidth]{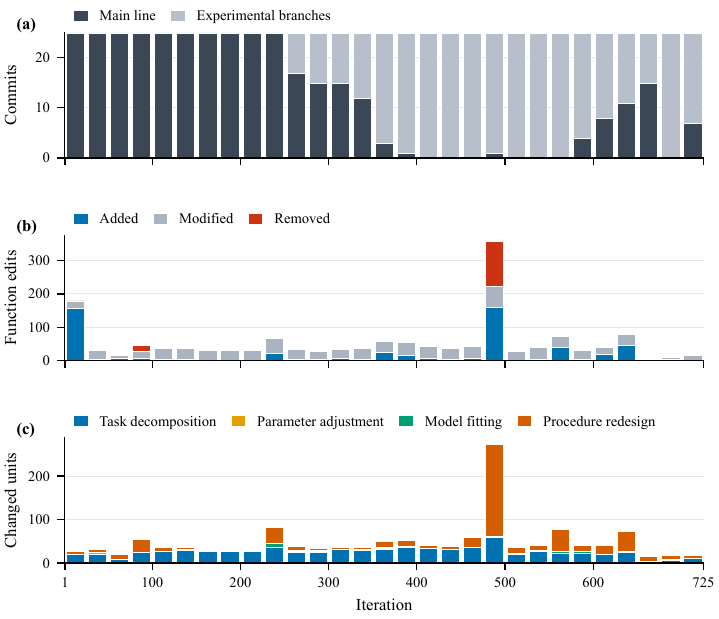}
\caption{Revision activity per 25 iterations. (a) Commits on the main line and on experimental branches. (b) Added, modified and removed workflow functions. (c) Changed code units by revision operation.}
\label{fig:iter-activity}
\end{figure}

\paragraph{Where the code changes.}
Revision work moves from the main line to experimental branches and back (Figure~\ref{fig:iter-activity}a). The first 258 iterations all lie on the main line. From iteration 259 onward, most work happens on experimental branches, and selected changes later return to the main line. Edits arrive in two large bursts (Figure~\ref{fig:iter-activity}b). The first iteration builds the initial workflow. Iterations 486--496 restructure Negotiation: one commit splits its tactics into separate modules, and another refactors them without a policy change. Between these bursts, most edits modify existing functions. Task decomposition appears in nearly every interval, while procedure redesign comes in bursts (Figure~\ref{fig:iter-activity}c). Decomposition changes 738 units in 458 commits, and redesign changes 591 units in 188 commits. Parameter adjustment and model fitting each change fewer than 35 units. Most revision work thus sharpens the conditions of existing subtasks.

\begin{figure}[tbp]
\centering
\includegraphics[width=\textwidth]{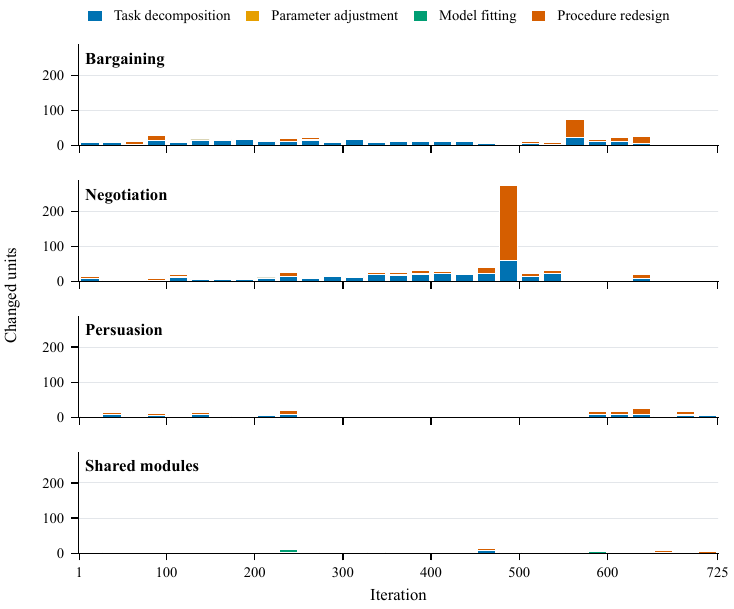}
\caption{Changed code units per 25 iterations by game and revision operation. Shared modules hold code and learned artifacts used by more than one game.}
\label{fig:iter-families}
\end{figure}

\paragraph{Different games in different periods.}
Each game is revised in its own period (Figure~\ref{fig:iter-families}). Bargaining changes throughout development. Negotiation work grows between iterations 300 and 550 and contains the restructuring burst. Persuasion changes in the first 250 iterations and again after iteration 580. Model fitting first appears at iterations 240--243, inside the interval where all three ratings rise in the main text. This timing does not show that fitting causes the rise.

\begin{figure}[tbp]
\centering
\includegraphics[width=\textwidth]{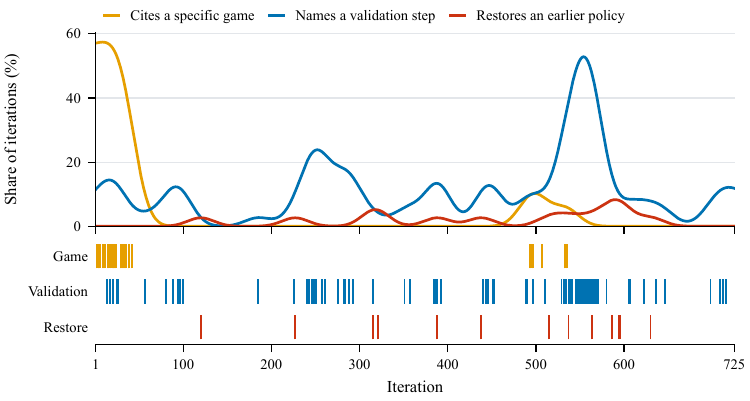}
\caption{Evidence named in commit messages across iterations. Curves give the smoothed share of iterations with each mark. Ticks below mark the individual commits.}
\label{fig:iter-evidence}
\end{figure}

\paragraph{How the coding agent justifies and undoes changes.}
The coding agent changes how it argues for a revision (Figure~\ref{fig:iter-evidence}). In the first 50 iterations, many commit messages cite a specific game and patch the case it reveals. One message records a game that rejected a 44\% share in round one and accepted the same share in round seven. The next workflow accepts 43\% or more through round six. Such citations then nearly vanish, and only a few later Negotiation fixes cite games again. Later messages name a validation step, such as a replay or a shadow run. These mentions peak near iterations 250 and 550. Thirteen commits restore an earlier strategy policy, and most of them come after iteration 300.

Some policies change direction several times before they settle (Table~\ref{tab:iter-backforth}). The Persuasion seller first skips a low-quality product, then recommends it and then skips it again. At iteration 31 the agent ends this swing with a new condition. The seller recommends low-quality products until the first sale and changes behavior after it. The disagreement is thus resolved by decomposition. A Bargaining plateau exit runs in shadow mode, goes live, is reverted and returns with a guard. It is then split by horizon type. A Rubinstein responder is promoted and demoted twice within eight iterations. Evolution therefore includes trial, withdrawal and repair besides steady growth.

\begin{table}[tbp]
\centering
\caption{Policies that change direction during development. Each step gives the iteration and the behavior that follows it.}
\label{tab:iter-backforth}
\begin{tabular}{@{}p{0.27\textwidth}p{0.69\textwidth}@{}}
\toprule
Policy & Steps \\
\midrule
Persuasion seller with a low-quality product & 1 skip $\rightarrow$ 22 recommend $\rightarrow$ 29 skip $\rightarrow$ 31 recommend until the first sale \\
Bargaining plateau exit & 548 shadow $\rightarrow$ 552 live $\rightarrow$ 564 reverted $\rightarrow$ 579 live with a guard $\rightarrow$ 585 split into finite and open horizons \\
Bargaining Rubinstein responder & 553 shadow $\rightarrow$ 556 live $\rightarrow$ 557 shadow $\rightarrow$ 559 live on a narrower band $\rightarrow$ 560 shadow \\
\bottomrule
\end{tabular}
\end{table}

\begin{figure}[tbp]
\centering
\includegraphics[width=\textwidth]{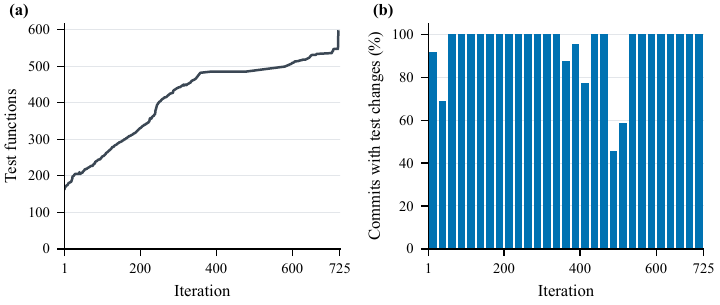}
\caption{Regression tests during development. (a) Test functions on the main line. (b) Share of strategy commits per 25 iterations that also change tests.}
\label{fig:iter-tests}
\end{figure}

\paragraph{How changes are checked.}
Revisions carry regression tests (Figure~\ref{fig:iter-tests}). Test functions on the main line grow from 165 to 595. Of the 527 commits that change strategy code, 493 also change tests. The share drops only around the restructuring burst. Each test fixes a diagnosed problem as an expected decision (Table~\ref{tab:iter-cases}). Later revisions can then change one branch and still detect changes in the others.

\begin{table}[tbp]
\centering
\caption{Diagnosed problems, the revisions they trigger and the regression tests that record them.}
\label{tab:iter-cases}
\begin{tabular}{@{}rp{0.2\textwidth}p{0.22\textwidth}p{0.4\textwidth}@{}}
\toprule
Iteration & Problem & Revision & Test checks that \\
\midrule
97 & Message wording misread & Refine conditions & a current recommendation outweighs earlier skip wording \\
110 & Opponent style misjudged & Update the estimate & recent stalls, not an early concession, set the style \\
285 & Low-payoff deals accepted & Refine conditions & a non-final offer at the bottom of the field is not accepted \\
608 & Same message rejected again and again & Refine conditions and change the procedure & a rejected counter message is not sent again \\
676 & Opponent misjudged & Update the estimate and refine conditions & the current opponent's revealed outcomes override the global prior \\
\bottomrule
\end{tabular}
\end{table}

\paragraph{Limits.}
The operation labels come from syntactic rules. A changed condition may fix a bug without splitting a subtask, and a mixed change counts once for each operation it contains. The counts measure code change and not its effect on the rating. Not every experimental branch was deployed. Commit messages show how the agent describes a change. They do not prove that the described validation ran.

%% file: figures/observed_refinement.tex
\begin{tikzpicture}[
  x=1cm,y=1cm,>=Latex,
  box/.style={draw=black!45,rounded corners=1pt,fill=black!2,
    align=center,font=\scriptsize,inner sep=4pt},
  test/.style={box,fill=blue!4,text width=2.4cm},
  terminalbox/.style={box,fill=green!7,text width=1.55cm},
  edge/.style={->,draw=black!65,line width=.55pt},
  note/.style={font=\scriptsize,align=left}]
\node[font=\small\bfseries,anchor=west] at (0,0) {(a) Branch added at commit 236};
\node[test] (entry) at (2,-.65) {Earlier acceptance conditions do not apply};
\node[test] (pat) at (2,-1.7) {Unknown horizon?\\$\delta_{\rm self},\widehat\delta_{\rm opp}\geq .99$?};
\node[test] (repeat) at (2,-2.85) {$\geq4$ repeated offers?\\$\geq4$ rejected\\own offers?};
\node[test] (share) at (2,-4) {Current self-share\\$0.45\leq s<0.48$?};
\node[terminalbox] (accept) at (2,-5) {Accept};
\node[box,text width=1.65cm] (old) at (4.65,-2.85) {Continue existing decision procedure};
\draw[edge] (entry)--(pat);
\draw[edge] (pat)--node[left,font=\scriptsize]{yes}(repeat);
\draw[edge] (repeat)--node[left,font=\scriptsize]{yes}(share);
\draw[edge] (share)--node[left,font=\scriptsize]{yes}(accept);
\draw[edge] (pat.east)--++(.35,0)-|node[pos=.2,above,font=\scriptsize]{no}(old.north);
\draw[edge] (repeat.east)--node[above,font=\scriptsize]{no}(old.west);
\draw[edge] (share.east)--++(.35,0)-|node[pos=.2,below,font=\scriptsize]{no}(old.south);
\draw[black!18] (6,-.25)--(6,-5.3);
\node[font=\small\bfseries,anchor=west] at (6.35,0) {(b) Evidence behind the revision};
\node[note,text width=6.3cm,anchor=north west] at (6.35,-.5)
 {Problem: a patient pair repeatedly rejects the same split. The revision makes the observed plateau an explicit acceptance condition.};
\node[box,text width=6.0cm,anchor=north west] at (6.35,-1.5)
 {Our requests: $50\%$. Opponent offers: $45.2\%$.\\
 Earlier offers: all rejected. Current round: 8.\\
 Own $\delta=1.0$; hidden opponent information;\\
 unknown horizon.};
\node[note,text width=6.2cm,anchor=north west] at (6.35,-2.9)
 {Checked-in regression expectations};
\node[note,anchor=west] at (6.55,-3.45) {Repeated sub-half offer};
\node[font=\scriptsize,text=red!65!black] at (9.7,-3.45) {reject};
\node[font=\scriptsize,text=green!40!black] at (11.9,-3.45) {accept};
\draw[edge] (10.2,-3.45)--(11.25,-3.45);
\node[note,text width=6.1cm,anchor=north west] at (6.55,-3.95)
 {New fixture: $45.2\%$ is accepted. Changing the opponent's offers to $48\%$ yields rejection: the added branch is a specific exception, not a general relaxation.};
\node[note,text width=6.1cm,anchor=north west,text=black!65] at (6.55,-4.95)
 {Source: versioned policy and regression tests; not an estimate of competition-score gain.};
\end{tikzpicture}

%% file: 9ExtendedExperiments.tex
\section{Competition Results}
\label{app:results}
This section adds per-game detail to the ratings in Table~\ref{tab:competition}. It reports the fixed-workflow samples behind the behavior analysis and the strategy each workflow uses. It then locates the large rating changes.

Table~\ref{tab:competition-details} reports the September~1 competition snapshot. Figure~\ref{fig:experiments} places the deployment-rating history and documented strategy changes on the same commit axis. A retained Bargaining deployment records the final 2760.27 rating; final Negotiation and Persuasion snapshot values remain separate from the commit-mapped curves.

\begin{table}[htbp]
\centering
\caption{Competition results, fixed-workflow samples, and examined strategies. The September~1 snapshot contains 42,917 cumulative games across the three environments. Fixed-workflow statistics describe a separate analysis sample of 1,112 games and 12,249 action records, including messages; they are not additional games to add to the competition total. Ratings are snapshot values, not averages over the fixed batches.}
\label{tab:competition-details}
\small
\begin{tabular*}{\textwidth}{@{\extracolsep{\fill}}lrrr@{}}
\toprule
Measure & Bargaining & Negotiation & Persuasion \\
\midrule
\multicolumn{4}{@{}l}{\textit{Competition snapshot}} \\
Official rating $\uparrow$ & 2760.27 & 2761.67 & 2261.58 \\
Cumulative games & 8,256 & 27,226 & 7,435 \\
\addlinespace
\multicolumn{4}{@{}l}{\textit{Fixed-workflow behavior sample}} \\
Games with action records & 512 & 199 & 401 \\
Recorded actions & 2,344 & 1,912 & 7,993 \\
Mean recorded actions/game & 4.58 & 9.61 & 19.93 \\
Rule-labeled actions & 2,022 & 16 & 4,987 \\
Numerical actions & 283 & 1,887--1,896 & 3,006 \\
Predictor actions & 8 & 0--9 & 0 \\
Unclassified actions & 31 & 0 & 0 \\
Recorded valid LLM selections & 0 & 0 & 0 \\
Runtime LLM judging & Off & Off & Off \\
LLM-generated messages & Off & Off & Off \\
Policy updates during deployment & Off & Off & Off \\
\bottomrule
\end{tabular*}

\medskip
\begin{tabular}{@{}p{.17\textwidth}p{.79\textwidth}@{}}
\toprule
Environment & Examined workflow strategy \\
\midrule
Bargaining &
Route by horizon, discounting, and opponent behavior. Construct offers from named plans; compare incoming shares with reservation and continuation values. Opening imitation and low-share prediction apply under explicit conditions, with earlier branches taking precedence. \\
\addlinespace
Negotiation &
Compute own profit, distinguish complete from incomplete information, and select an offer or acceptance procedure. Estimate the opponent's acceptance boundary, adjust concessions from observed offers, and retain opponent profiles across interactions where configured. \\
\addlinespace
Persuasion &
Condition recommendations on quality, previous sales, and buyer responses. Update buyer beliefs from revealed outcomes and choose purchases by expected value; seller procedures adapt signals and message templates to the observed interaction history. \\
\bottomrule
\end{tabular}

\smallskip
\begin{minipage}{\textwidth}
\small
Action labels describe the final recorded procedure, not all upstream computations, and use environment-specific definitions. Nine Negotiation actions are ambiguous between numerical and fitted selection, giving the coupled bounds above; the bounds must not be added independently. Policy freezing fixes deployed strategy parameters but does not prevent conditioning on the current interaction history. Strategy descriptions summarize mechanisms examined in the code and traces, not components active in every competition game.
\end{minipage}
\end{table}

\subsection{Large Rating Changes}
The recorded interval ending at position 243 contains increases of 107.16 rating points over 267 Bargaining games, 147.19 over 262 Negotiation games, and 139.64 over 267 Persuasion games. Thus, the visible increase is not confined to the Bargaining branch annotated on the curve. The extended history captures later changes: the Bargaining rating rises from 2331.96 to 2631.70 at the same recorded revision, then reaches 2760.27 at a subsequent deployment. Negotiation declines from 2923.32 to 1844.08 before recovering. These are changes between rating observations, not single-match effects or controlled estimates of a code edit. Appendix~\ref{app:revision} therefore examines the implementation and behavior records to explain what changed in the strategy itself.

\section{Hybrid Decision Making}
\label{app:hybrid}
The agent decides through code, fitted models and LLM calls; this section measures how much of the play each one decides. It first counts decision points in the source code and then attributes recorded actions to executors. It ends with how often the LLM is called and what that use is associated with.

\subsection{Decision Sites in the Source Code}
The module shares in Figure~\ref{fig:decision-profile} come solely from committed source code. They do not use action logs or infer an executor from missing log entries. We inspect the three environment-specific workflow directories at each of the 359 main-branch revisions. A code site is one syntactic conditional statement, conditional expression, while-condition, or match selector. Compound Boolean tests count as one site rather than one site per comparison. The inventory includes intermediate routing and eligibility checks as well as final-action checks; it is not restricted to leaves.

An ML site is an explicit call to the fitted prediction interfaces, segmented prediction interfaces, learned-policy lookup, or rating-surrogate prediction interface. Imported aliases are resolved before classification. Model-internal arithmetic and coefficient-loading routines are not counted again. The shared LLM candidate-selection interface contributes one site per environment where the player calls the judge and the source does not unconditionally reject that environment. Runtime configuration may disable this interface; its presence does not imply an actual invocation. Message generation and the agent's workflow-editing calls are outside this decision-site inventory.

We sum site counts across the three environments and divide each executor count by their sum. These percentages partition the defined sites, not executed actions. Source definitions can contain dormant functions or guarded paths; this inventory does not prove path feasibility. Counts depend on the chosen syntactic granularity, so a model interface and its internal calculations must not be interpreted as equal computational work. File paths, blob hashes, line numbers, expressions, and revision-to-source mappings accompany the exported counts. There is no smoothing or interpolation between sampled revisions; the curves use steps.

\subsection{Executor Attribution of Actions}
The separate runtime-attribution analysis counts executed non-message actions, not functions, distinct code branches, or construction-time agent calls. An LLM label denotes a persisted valid candidate selection, including selections that retain the original action. Code combines the audited rule and numerical procedures. An ML-model label denotes an identified fitted-model decision. Records without sufficient evidence remain unclassified and stay in the denominator. These runtime statistics are not the source of Figure~\ref{fig:decision-profile}.

The historical panel reuses the 57,179 replay actions and 233 valid-selection records described below. The remainder is unclassified because the replay export lacks exhaustive executor labels; a zero confirmed count for code or ML does not establish zero use. The trailing window spans ten main-branch commit positions, with numerator and denominator summed before normalization. Positions are mapped onto the shared commit index, and intervals without recorded actions are missing rather than zero. This panel therefore shows attribution coverage and confirmed decisions, not a complete three-executor transition.

The fixed-version panel excludes 1,957 seller-message records from the 12,249-row sample, leaving 10,292 non-message actions. Bargaining contains 2,305 code decisions, eight identified predictor decisions, and 31 unresolved decisions. Negotiation contains 1,903 code decisions and nine unresolved decisions for which a fitted candidate is recorded but its selection over the numerical fallback is not identifiable. All 6,036 non-message Persuasion actions use the audited code procedures. Runtime LLM selection is disabled in these three samples. Their observed executor shares are not extrapolated to earlier versions, and the Persuasion sample is not assigned a historical commit because its loaded runtime hash is unknown.

\subsection{Condition-to-Action Checks}
We test alignment with the implemented policy by checking branch conditions against recorded actions. During development, regression tests accompany nearly every strategy revision (Appendix~\ref{app:iteration-activity}). In the fixed Bargaining batch, opening imitation selects a $55\%$ self-share in all 63 eligible states. The reservation condition applies to 451 states, but its rejection procedure handles only 382: the remaining 69 take higher-precedence branches, producing 57 rejections and 12 acceptances. The low-share predictor condition activates eight times, all followed by rejection (Table~\ref{tab:behavior}).

\begin{table}[htbp]
\centering
\caption{Condition-to-action checks on fixed Bargaining records. Eligibility does not imply execution when an earlier condition takes precedence.}
\label{tab:behavior}
\small
\begin{tabular}{@{}lrrl@{}}
\toprule
Condition & Eligible & Executed & Recorded action \\
\midrule
Opening imitation & 63 & 63 & offer $55\%$ self-share \\
Reservation lock & 451 & 382 & reject below $60\%/47\%$ \\
Low-share predictor gate & 8 & 8 & reject \\
\bottomrule
\end{tabular}
\end{table}

The inventory contains 512 Bargaining games with 2,344 action rows, 199 Negotiation games with 1,912 rows, and 401 Persuasion games with 7,993 rows. Every trace matches the target agent, assigned batch, game family, and deployment identifier. All three deployments set \code{freeze\_policy=true}, \code{llm\_judge=false}, and \code{llm\_messages=false}. The batch outcome ledgers are partial or contaminated, so we use their action traces but not batch rating deltas.

The exporter checks Bargaining routes directly from persisted state, actions, and named selection nodes. It finds 63 states eligible for the field-imitation opening, all 63 routed to the node, spanning 60 games with no action violation. The raw patient-reservation predicate holds for 451 states: 382 route to the reservation lock, while 69 take higher-priority veto or exit branches. The reservation activations span 83 games and have no branch violation. The severe-low-share predictor gate appears eight times in six games, and all eight actions reject. A separate legality pass finds zero schema or offer-budget violations among all 2,344 Bargaining action rows from 512 games. These checks establish behavior on logged states, not counterfactual route coverage or the accuracy of every upstream semantic classification.

\subsection{Executor Labels in Frozen Deployments}
\begin{figure}[H]
\centering
\includegraphics[width=0.82\textwidth]{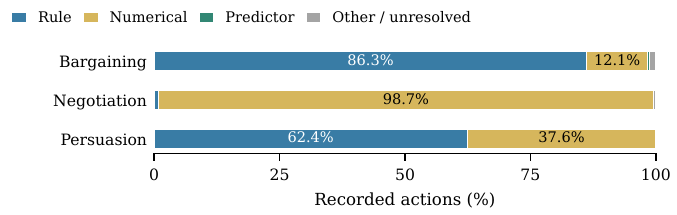}
\caption{Final decision labels in fixed-workflow records. Grey segments include 31 unclassified Bargaining records and nine unresolved Negotiation records; eight Bargaining records use a predictor. Runtime LLM judging is disabled in all three deployments. Label definitions are environment-specific; detailed counts and bounds are given in the text.}
\label{fig:executors}
\end{figure}

The checked conditions match recorded behavior, and all 2,344 Bargaining actions satisfy schema and budget constraints; this does not establish decision optimality. The fixed batches contain 512 Bargaining, 199 Negotiation, and 401 Persuasion games. Figure~\ref{fig:executors} shows their recorded decision labels. Bargaining has 2,022 rule-labeled actions, 283 numerical actions, eight predictor actions, and 31 unclassified actions. Negotiation has 16 rule-labeled actions, 1,887--1,896 numerical actions, and up to nine predictor actions; those nine records do not identify whether the fitted candidate displaced its fallback. Persuasion has 4,987 rule-labeled and 3,006 numerical actions. These labels identify the final recorded procedure, not every computation performed earlier in the workflow. Their definitions differ across environments, preventing direct comparison as a common measure of model dependence.

For Bargaining, the severe-low-share fitted node is a predictor; trajectory, model, strict-dominance, and risk stages are numeric; and legacy semantic stages are rules. This yields 2,022 rule, 283 numeric, and 8 predictor rows, while 31 rows in four newer stages remain unclassified. For Negotiation, \code{walk} is a rule and other named procedures are numeric by default. Nine \code{faratin\_boulware} rows expose paired fitted scores but do not record whether a fitted proposal displaced its numeric fallback, producing only loose bounds of 16 rule, 1,887--1,896 numeric, and 0--9 predictor rows. Persuasion's expected-value procedures are numeric and its trust-, history-, or threshold-conditioned procedures are rules; source inspection classifies three \code{learned\_low\_text} rows as rules, producing 4,987 rule and 3,006 numeric rows. These categories describe final recorded selection nodes, not every upstream computation. The family mappings are not equivalent: Bargaining uses final-node/stage labels, whereas Negotiation and Persuasion use named-procedure taxonomies. Persuasion also lacks a runtime commit, so its action evidence is deployment-specific but not commit-aligned.

\subsection{LLM Judge Usage}
\begin{figure}[H]
\centering
\includegraphics[width=\textwidth]{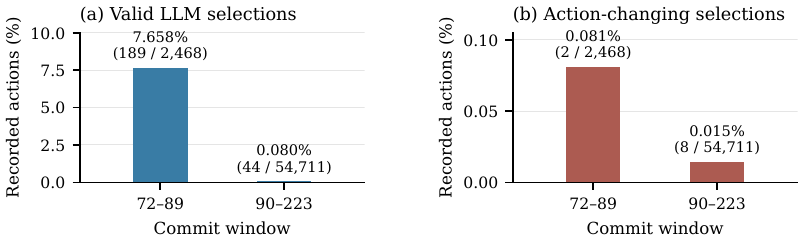}
\caption{Recorded LLM decisions across two commit windows. (a) Valid model selections as a percentage of recorded non-message actions. (b) Selections that change the program's proposed action, using the same denominator. Labels include raw counts. Both axes are linear but use different ranges; the plotted events do not count every attempted call.}
\label{fig:runtime}
\end{figure}

A bounded judge appears at position 40, is restricted to ambiguous cases at 55, and is disabled for Bargaining at 186. Windows 72--89 and 90--223 contain 189/2,468 and 44/54,711 valid picks per non-message action; only 2 and 8 change the action (Figure~\ref{fig:runtime}). These correspond to valid-pick rates of $7.66\%$ and $0.080\%$, while action-changing picks account for $0.081\%$ and $0.015\%$ of recorded actions. The later window contains fewer observed interventions relative to its action count, but coverage differs between windows. Failed or unrecognized invocations are missing, precluding a total inference-cost estimate.

Judge logs linked to games cover iterations 72--223. In iterations 72--82, the judge answers 9.1\% of non-message actions, and 185 of its 189 responses are Persuasion buyer decisions. It agrees with the code in 184 of these 185 cases. Iterations 83--84 and 90--91 make these decisions deterministic. The judge then answers 0.08\% of actions in iterations 90--223, and ten responses in iterations 72--223 change the chosen action.

We reuse the episode-to-deployment alignment and count non-message action records, persisted valid judge selections, and action-changing selections separately at each main-branch commit position from 72 to 223. Their totals are 57,179, 233, and 10. The archived logged-decision analysis uses trailing windows ending at each commit: numerators and denominators are summed before computing percentages, rather than averaging per-commit rates. The window includes at most ten commit positions and is truncated at position 72 initially. A window with no recorded actions is missing, not a zero rate. Raw window action counts are retained in the derived CSV. These are descriptive overlapping-window statistics, not independent trials or estimates of total model calls; action counts are observation coverage, not compute cost. Main-branch positions differ from the cross-branch index in Figure~\ref{fig:strategy-timeline}.

\subsection{Runtime Parameters and Timing}
Runtime LLM calls use temperature 0.2 and a 120-token output limit. Decision times average ten replays of each logged state with warmed code. Workflow statistics in Table~\ref{tab:competition} count named classifiers as multiway nodes and exclude technical checks. Figure~\ref{fig:decision-profile}a smooths module shares over 11 iterations. Figure~\ref{fig:decision-profile}b times 1,420 code and 450 regression attempts and five LLM calls. Two LLM calls return valid responses and three fail. Table~\ref{tab:competition} averages ten timings per state over 67, 60 and 60 offline states.

\subsection{Model Use and Rating Changes}
The score panel uses disjoint 20-commit intervals beginning at position 72, with the last interval ending at 223. Within each family and interval, we retain unique replay games with recorded non-message actions and finite rating deltas. The rate numerator, action denominator, and mean game-level rating change all use these same games. The finite-rating subset contains 5,477 games and 57,097 actions; five games with 82 action records are excluded because a finite rating delta is unavailable. Each point is one family--interval group; groups need at least ten games, and all 24 groups meet this threshold. No score shift is attributed to individual model decisions: versions, opponents, and game conditions differ, and no counterfactual execution was performed. Model-log absence does not establish code execution. Historical turn exports contain actions and outcomes but no exhaustive final-executor labels, so they do not support a three-way LLM/code/predictor area chart. The fixed-deployment taxonomy in Table~\ref{tab:competition} is not extrapolated to earlier commits.

\section{Logs and Commit Histories}
\label{app:logs}
All counts in this paper come from deployment logs and the commit histories of the game repository. This section lists these data and explains how the commit counts relate. It then describes how episodes are aligned with deployments and what cannot be released.

\subsection{Data Sources}
We study three GLEE environments across 359 first-parent commits (\emph{commit positions}). We align 166 startup-rating records with 159 revisions, ending at position 256. Three fixed-workflow deployments provide 1,112 games and 12,249 action records, including messages, with runtime LLM judging disabled. Archived deployment logs extend the performance history to 799 family-rating observations across 389 recorded revisions. Figure~\ref{fig:strategy-timeline} shows a segmented summary, retaining one observation per plotted revision. Deployment hashes locate observations; they do not establish which edit caused a rating change.

\subsection{Commit Indices}
The paper uses two commit indices. The main branch's first-parent path has 359 commits. Its positions serve the deployment alignment and the case studies. The shared index merges these commits with family-lineage branches and additional deployment commits. It deduplicates hashes and orders the 725 commits by timestamp. The figures and the source audit use this index. Of its 725 commits, 530 touch strategy code.

\subsection{Episode-to-Deployment Alignment}
We enumerate 359 commits on the main branch's first-parent path. Each episode is assigned to the latest target-agent deployment whose timestamp does not exceed its start time, and that revision is mapped to a first-parent position. Judge logs begin carrying game IDs at position 71. Positions 72--89 include 330 episodes, of which 301 have 2,468 recorded non-message actions; positions 90--223 include 5,672 episodes, of which 5,181 have 54,711. The turn export skips 25,216 empty or message-only actions. We count a valid pick when a matched judge outcome exists and an action-changing pick only when that event replaces the deterministic candidate. Failed, timed-out, cooldown, or unrecognized picks return before this event and usually lack a game ID, so invocation totals cannot be aligned to deployment windows. The pre-Git agent log records 586 direct-LLM actions but lacks a turn denominator and commit mapping.

The deployment mapping ends at commit position 256. Positions 257--315 are checked-in code milestones without a retained target-deployment assignment, and the verified September 1 rating endpoint is likewise commit-unassigned.

\subsection{Release Limits}
The local research package contains exporter scripts and the derived CSV/JSON used for the figures. It is not yet a public reproduction artifact: raw interaction logs cannot be redistributed, and the exporter contains local deployment identifiers. Without those logs, the historical export cannot be regenerated end to end.

\section{Compliance with GLEE Rules}
\label{app:compliance}
GLEE permits hand-coded heuristics and agent-code updates. The workflow uses only API-provided game state, visible history, and legal-action metadata. It does not automate the Human Track, select actions through live human intervention, coordinate accounts, or exploit undocumented game behavior.